\documentclass{article}

\usepackage[final,nonatbib]{neurips_2023}

\usepackage[utf8]{inputenc} 
\usepackage[T1]{fontenc}    
\usepackage{hyperref}       
\usepackage{url}            
\usepackage{booktabs}       
\usepackage{amsfonts}       
\usepackage{nicefrac}       
\usepackage{microtype}      
\usepackage{xcolor}         
\usepackage{multirow}
\usepackage{graphicx}
\usepackage{subcaption}
\usepackage{amsmath}

\title{Strong statistical parity through fair synthetic data}

\author{%
  Ivona Krchova\\
  MOSTLY AI\\
  \texttt{ivona.krchova@mostly.ai} \\
  \And
  Michael Platzer \\
  MOSTLY AI \\
  \texttt{michael.platzer@mostly.ai} \\
  \AND
  Paul Tiwald \\
  MOSTLY AI \\
  \texttt{paul.tiwald@mostly.ai} \\
}

\begin{document}

\maketitle

\begin{abstract}
  AI-generated synthetic data, in addition to protecting the privacy of original data sets, allows users and data consumers to tailor data to their needs. This paper explores the creation of synthetic data that embodies Fairness by Design, focusing on the statistical parity fairness definition. By equalizing the learned target probability distributions of the synthetic data generator across sensitive attributes, a downstream model trained on such synthetic data provides fair predictions across all thresholds, that is, strong fair predictions even when inferring from biased, original data. This fairness adjustment can be either directly integrated into the sampling process of a synthetic generator or added as a post-processing step. The flexibility allows data consumers to create fair synthetic data and fine-tune the trade-off between accuracy and fairness without any previous assumptions on the data or re-training the synthetic data generator.
\end{abstract}

\section{Introduction}

In recent years, the advent of privacy-preserving, AI-generated synthetic data, which we will refer to as synthetic data in the following, has brought a revolutionary change to data privacy. Privacy-preserving synthetic data offers Privacy by Design by strongly mitigating data privacy risks for data consumers. Once a data set is replaced with its privacy-preserving synthetic version, privacy risks are mitigated for all subsequent downstream applications. However, the utility of synthetic data extends far beyond privacy. It enables data consumers to fine-tune data sets to meet their needs, making it a highly adaptable tool. In this paper, we explore the ability of synthetic data to provide not only Privacy by Design but also Fairness by Design.\\
The process of generating synthetic data involves training generative models on the original data sets, resulting in the creation of entirely new records that faithfully reflect the statistical patterns contained in the source data. When the original data set contains biases, the synthetic data inadvertently maintains these biases. This phenomenon is not unique to synthetic data but applies to any algorithm as machine learning models tend to inherit biases from their training data.\\
Responsible AI strives to design, develop, and deploy artificial intelligence systems in a way that, ideally, everyone profits and no one is harmed. One of the pillars of responsible AI is fairness which ensures that the algorithm should not discriminate against people because of their race, gender, age, socioeconomic status, and other, so-called, sensitive attributes.\\
In this work, we focus on the generation of synthetic data adhering to the statistical parity fairness definition, as statistical parity has a direct relationship to the data and can be readily controlled during synthetic data generation, without the need to feed back information and performance metrics from any downstream tasks. We investigate how training a classifier on fair synthetic data affects its predictions on biased original data.\\
 The common approach in machine learning involves training a classifier while carefully tuning the decision threshold. This threshold separates the probabilities predicted by a machine-learning model into different classes. For example, in a binary classification, we have a predicted probability indicating the likelihood of a positive outcome for each individual record. If the predicted probability for a truly positive record exceeds the threshold, the instance is classified as positive; otherwise, it is classified as negative making it a “false negative”. Vice versa, if the predicted probability for a truly negative instance exceeds the threshold, it is classified as positive, making it a "false positive". When deploying a classifier, it is usually necessary to tune the threshold and evaluate whether the resulting model meets the domain- and business-specific needs. As these needs are often subject to change and differ from application to application, the primary objective of this research paper is to achieve fairness across all thresholds simultaneously.\\
Just as it is important to create synthetic data that is privacy-preserving, it is equally important, in the context of fairness, to be aware of the shortcomings of “naively fair synthetic data”. Naively fair synthetic data conform to the definitions of fairness (statistical parity) within the synthetic data itself but fail to propagate fairness to the final predictions of downstream classifiers trained on said naively fair synthetic data. In the context of statistical parity, naively fair data can be produced by simply re-sampling records from unprivileged groups with favorable target outcomes to equalize the proportion of favorable outcomes across different sensitive groups. In contrast, truly fair synthetic data are carefully constructed to ensure fair predictions when used to train downstream classifiers.\\
Our work is based on techniques developed in fairness research, which are capable of producing predictions that are fair across all thresholds. In this paper, we extend these techniques to the creation of fair synthetic data and show the stability of this fair synthetic data to propagate fairness to downstream models. These threshold-independent fair predictions bring us a step closer to establishing Fairness by Design.


\section{Related work}
There are three different places in an ML pipeline where strategies for mitigating bias can be implemented \cite{fairsummary}. 
\textit{Pre-processing} strategies focus on modifying the data set before it is used to train machine learning models. The aim is to remove or mitigate any potential biases present in the data with respect to the protected attributes. \textit{In-processing} strategies aim to directly modify the learning algorithm to make it fair during the training process. These methods try to optimize fairness as part of the model's training objective. \textit{Post-processing} strategies modify the model's predictions/outputs after it has been trained to ensure fairness without changing the original data or the training process.\\
Considering a complete ML pipeline, fair synthetic data generation falls in the \textit{pre-processing} category as it changes the data used for training the final downstream model. For the remainder of this paper, we focus on the generation of fair synthetic data alone. Pre-, in-, and post-processing steps are, therefore, considered relative to the synthetic data generator.\\
Pre-processing strategies typically identify and remove original biased samples \cite{fairgen} before they even enter the synthetic data generator. The majority of contributions to fair synthetic data generation focuses on in-processing strategies trying to modify the synthetic data generator algorithm/training in order to generate fair synthetic data \cite{fairgan, tabfairgan, ourfairness}. Many studies introduce fairness by adding one or multiple fairness constraints to the objective function governing the training of the synthetic data generator. However, in-processing approaches exhibit limitations in terms of flexibility. They require tailoring the synthetic generator training to one or many specific sensitive columns and one target column. Additionally, the weight of the fairness constraints in the loss function can only be modified before or during training the generator. Striking the right balance between accuracy and fairness requires multiple iterations of training the synthetic-data generator which consumes valuable time and resources.\\
Moreover, studies have demonstrated that generating synthetic data that adheres to fairness definitions does not always guarantee fairness in downstream models \cite{notnecessarily}. This is studied in detail in the DECAF approach \cite{decaf}, where the objective extends beyond generating fair synthetic data to ensuring fairness in downstream model predictions. The DECAF algorithm enables inference-time debiasing, where edges between features can be strategically removed to satisfy user-defined fairness requirements. Consequently, this approach is more adaptable and focused on preserving fairness in downstream model predictions. However, it does come with the requirement that the data-generating process must be represented by a user-defined and causally sufficient directed acyclical graph. Causality-based fairness is also discussed in \cite{prefair}.\\
Our contribution is two-fold:
\begin{enumerate}
    \item We make the generation of fair-synthetic data flexible. We present a post-processing algorithm that adjusts synthetic data, that is the probability distributions outputted by the synthetic data generator which allows data consumers to select sensitive and target columns, and tune the strength of the fairness correction without retraining the synthetic data generator and without representing the original data in a causally sufficient graph. 
    \item We ensure that downstream classifiers trained on fair synthetic data yield fair predictions across arbitrary thresholds even when inferring from biased original data.
\end{enumerate}
Our approach is rooted in findings in fairness research \cite{wasserstein, georepair, quantileregression, fairregression}, which advocate the necessity of guaranteeing fair model predictions across all possible thresholds. Moving to the post-processing stage of the synthetic data generation process and inducing fairness across all possible thresholds makes our approach flexible for data consumers and provides strong statistical parity for downstream model predictions.\\

\section{Concept}
Consider a feature space $X \subseteq \mathbb{R}^d$, a set of $K$ protected attributes denoted as $S = \{s_1, s_2, ..., s_K\}$, and a binary target variable $Y = \{0, 1\}$, where 1 signifies the positive class and 0 the negative class. The primary goal of a fair learning algorithm is to learn insights about the target variable Y in a way that the conditional probability $P(Y | X)$ closely approximates $P(Y | X, S)$.\\
This paper explores strong demographic parity, an algorithmic fairness concept that ensures equally favorable outcomes across different protected attributes. With the model's output $Z$, typically a classification score, and the decision threshold $t$, strong demographic parity is defined as
\[P(Z \geq t | S=s_i) = P(Z \geq t | S=s_j) \quad \text{for } i, j \in \{1, 2, \ldots, K\}, \quad \forall t \in Z.\]\\
We focus on a binary classification problem as the downstream task, where $0 \leq Z \leq 1$, and two protected groups: a privileged group ($S=s_1$) and an unprivileged group ($S=s_2$). The objective of strong demographic parity is to ensure that the unprivileged group receives the same favorable outcome ($Z \geq t$) as the privileged group for any arbitrary threshold value $t$.\\
Our aim is to achieve an equal positive rate (PR) for both the unprivileged and privileged groups, as defined by the chosen threshold. The final goal of a fair learning algorithm under the strong demographic parity condition is to maximize the true positive rate (TPR) while minimizing the false positive rate (FPR), given the condition that the PR remains the same for both sensitive groups.\\
Equal PR translates to having the same $P(Z \geq t)$, which is equivalent to having the same $1 - P(Z < t)$. In probability theory, $P(Z \leq t) = F_t$, is the cumulative distribution function that describes the probability distribution of the random variable $Z$. The objective of having the same probability distribution can be mathematically expressed as minimizing the Wasserstein distance \cite{wassersteindistance} between two probability distributions: $P(Z | X, S=s_1)$ and $P(Z | X, S=s_2)$. This concept of fairness is theoretically described in \cite{wasserstein}, \cite{projectionfairness}.\\
In our implementation, we approximately align the empirical conditional probability distribution of the unprivileged group $P(Y | X, S=s2)$ with the one of the privileged group $P(Y | X, S=s1)$. Both distributions are extracted either directly from the synthetic generator $G$ or an independent post-processing classifier $C$ trained on the synthetic data. We align both distributions by learning a linear function $f$ which transforms a set of $M=100$ equidistantly spaced quantiles of $P(Y | X, S=s2)$ to match the same set of quantiles of $P(Y | X, S=s1)$. $f$ is then applied to the probability $P(Y=1|X, S=s2)$ of each synthetic instance of the unprivileged group in the sampling stage of either $G$ or $C$ to yield the modified probabilities $P^*(Y=1|X, S=s2) = f(P(Y=1|X, S=s2)$.\\
We use the free version of the MOSTLY AI synthetic data generator \cite{mostly} to create synthetic data subjects and then train the post-processing classifier $C$, a LightGBM model, to predict the target variable $Y$ on the synthetic data. $f$ is trained on $P(Y | X, S=s_{j})$ of $C$ and then used to get the modified probabilities $P^*(Y=1|X, S=s2)$. Finally, we use the $P^*(Y=1|X, S=s2)$s for sampling \cite{sampling} in order to assign each synthetic data subject of the unprivileged group either the value $0$ or $1$ in the target variable. This fair synthetic data is then used in the downstream task.\\
In order to ensure that the fairness in the synthetic data is propagated to the downstream task predictions, two requirements are key: first, a synthetic data generator capable of preserving the distribution of the real data, and second, the capability of the downstream model to predict the target variable to a comparable extent as the synthetic data generator.\\
To assess the first assumption, one can compare the distributions of the original and synthetic data, especially the positive rates in the target variable for the privileged and unprivileged groups. The second assumption requires the use of robust prediction models capable of accurately describing the relationship between the target variable and other attributes, thus yielding a strong performance metric.\\
For balancing the strength of the fairness correction, we introduce the parameter $\lambda$ ranging from zero to one. We define a new set of sampling probabilities by forming the convex combination $P^\lambda(Y=1|X, S=s2) = \lambda P^*(Y=1|X, S=s2) + (1-\lambda) P(Y=1|X, S=s2)$. For $\lambda=0$, no fairness correction is applied while for $\lambda=1$ we fully match - within the limits of our numerical approach - the sampling probabilities of the unprivileged class to the ones of the privileged class, i.e. we apply the strongest possible fairness correction.

\section{Experiments}
\subsection{Data sets}
We select two publicly available data sets that are commonly used in fairness-aware machine 
learning \cite{ds_survey}. Both data sets are characterized by a substantial statistical disparity. Further, to reasonably examine statistical parity, we select data sets in which each unprivileged group in the holdout data set contains a sample size exceeding 1000 observations, given our 80:20 split of the data into training and holdout subsets.
\paragraph{The Adult data set} \cite{adult_ds} involves a binary classification task: determining whether an individual's annual income exceeds \$50,000 based on demographic attributes. It consists of 48,842 instances, each described by 15 attributes. Following the recommendation in \cite{ds_survey}, we omit the \textit{fnlwgt} attribute. Additionally, we excluded the \textit{education-num} column, as it represents a one-to-one mapping of the \textit{education} attribute, expressed in numerical form. The class attribute is \textit{income} = \{$\leq$ \textit{50K}, $>$ \textit{50K}\}, the positive class is labeled as \textit{>50K}, and the sensitive attribute is gender, \textit{sex} = \{\textit{Male}, \textit{Female}\}. 30\% of males fall into the high-income category, while only 11\% of females do.
\paragraph{The Dutch census data set} \cite{dutch_ds} contains information about people in the Netherlands for the year 2001. It comprises 60,420 instances, each characterized by 12 attributes. This data set presents a binary classification task: predicting a person's occupation as either a high-level (prestigious) or low-level profession. The positive class is a high-level profession. The protected attribute here is also gender \textit{sex} = \{\textit{Male}, \textit{Female}\}. Nearly 63\% of males are associated with prestigious professions, while only 33\% of females hold such positions.

\subsection{Downstream models}
We use AutoGluon \cite{autogluon}, an open-source AutoML toolkit known for its robust predictive performance across various machine-learning models. Specifically, we leverage CatBoost, LightGBM, RandomForest, and XGBoost models using their default parameters. Our training process involves utilizing 80\% of the data set, which is also used for training the synthetic data generator model. The remaining 20\% of the data set is used as the holdout data set for model evaluation. We synthesize the training data set 10 times and we report the mean values as well as the standard deviations of selected metrics for each considered data set.

\subsection{Metrics and parameters}
\paragraph{Fairness:}
In assessing fairness, we systematically consider thresholds $t$ ranging from $0$ to $1$, with increments of $0.01$. We measure fairness using the statistical parity difference ($SPD$), which is defined as $SPD = P(Z \geq t \,|\, S=s_1) - P(Z \geq t \,|\, S=s_2)$, where $s_1$ signifies the privileged group (males), and $s_2$ represents the unprivileged group (females). We report means of absolute $SPD$ over all 101 thresholds and 10 synthetic runs along with their standard deviations.

\paragraph{Model performance:}
In general, the introduction of the fairness constraint leads to a decrease in model performance compared to scenarios without this additional correction. The extent of this performance drop is also dependent on the chosen evaluation metric. In our experiments, we use the area under the ROC curve ($AUC$) as our primary performance metric, due to its threshold-independent nature. We report means of $AUC$ over 10 synthetic runs along with their standard deviations.

\paragraph{$\lambda$ parameter:}
We report results for five distinct values of $\lambda$: 0, 0.25, 0.5, 0.75 and 1. Results are denoted as \textit{synfair\_$\lambda$}, where \textit{synfair\_0.0} and \textit{synfair\_1.0} showcase results without any and full fairness correction, respectively.

\subsection{Results}
\begin{figure}
    \centering
    \includegraphics[width=\columnwidth]{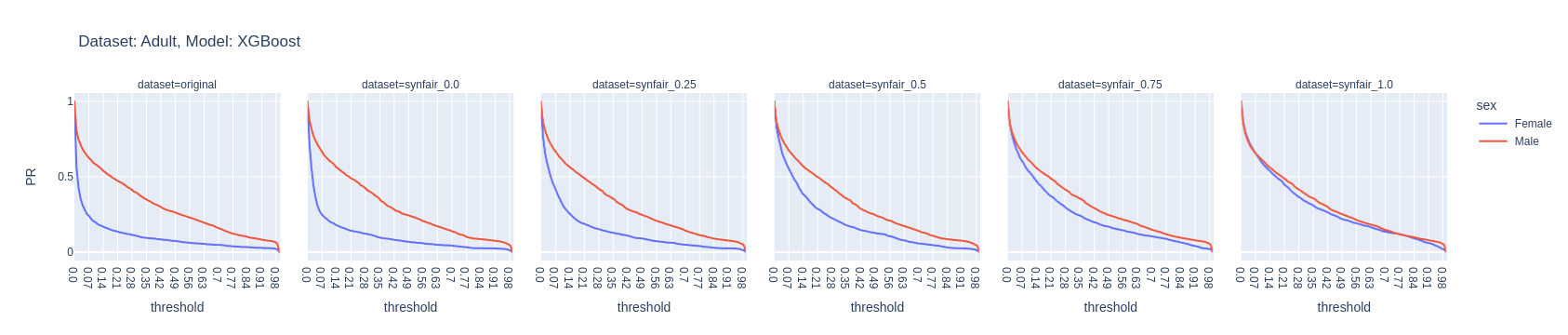}
    \includegraphics[width=\columnwidth]{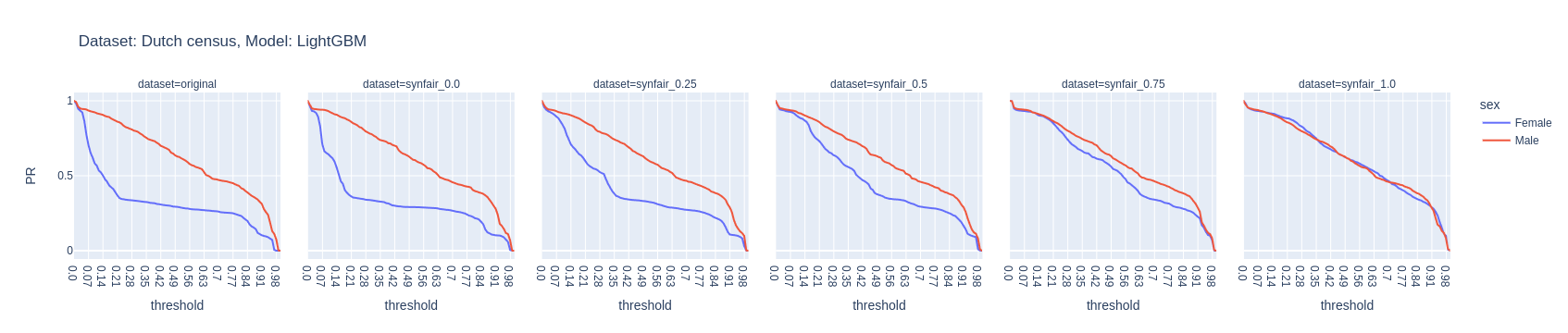}
    \caption{Positive rates across thresholds for each sensitive group. For the Adult data set, the PR of the XGBoost predictions is presented. For the Dutch census data set, the PR of LightGBM predictions is presented. As $\lambda$ increases, female positive rates approach male positive rates.}
    ~\label{fig:PR}
\end{figure}

\begin{figure}
    \centering
    \includegraphics[width=\columnwidth]{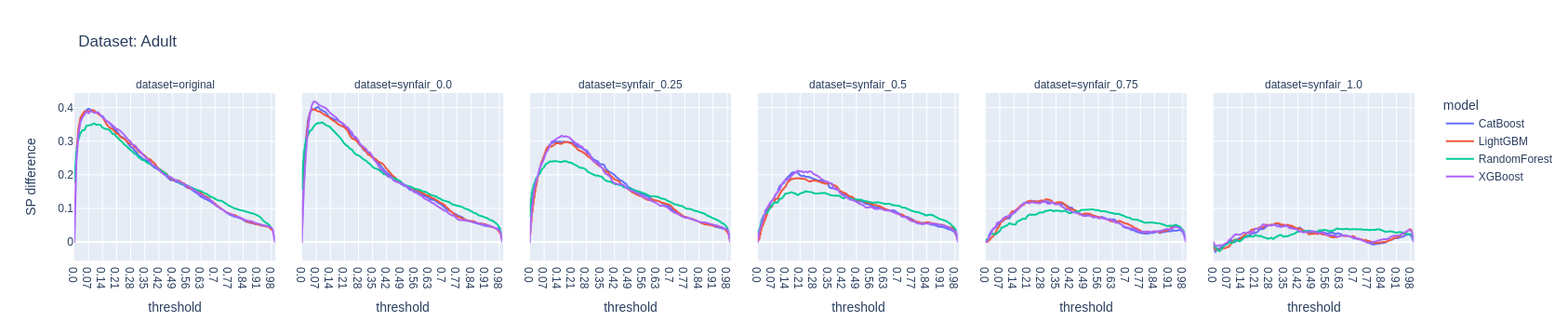}
    \includegraphics[width=\columnwidth]{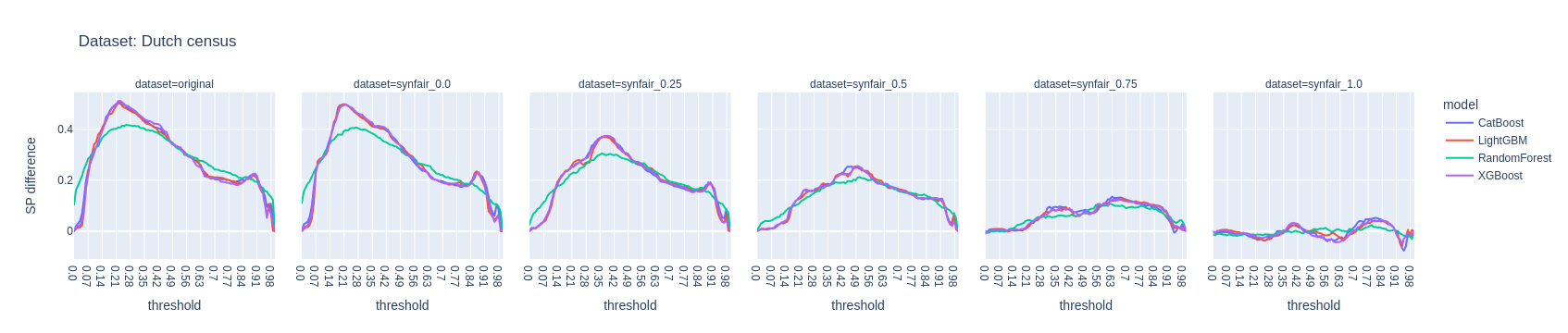}
    \caption{Statistical parity difference across thresholds for each downstream model and each training data. As $\lambda$ increases, the statistical parity difference decreases.}
    ~\label{fig:SPD}
\end{figure}

For both data sets in the \textit{synfair\_1.0} case, we successfully reduce the initial $SPD$ to a level well below 0.1 across all models (tables \ref{adult-table}, \ref{dutch-table}). While achieving statistical parity comes at the cost of overall accuracy, it is worth noting that this drop is limited to 5 percentage points and that it is adjustable through the $\lambda$ parameter. If it is not imperative to achieve near-perfect fairness but rather to simply reduce the disparity by half, selecting a lambda value of 0.5 allows for a smaller decrease in accuracy.\\
We also achieve a significant reduction of the initial $SPD$ across all thresholds (fig. \ref{fig:SPD}). Given the similarity of the results across all models and runs, we present positive rates graphically only for one run. For the best performing models, XGBoost and LightGBM on the Adult and the Dutch census data sets, respectively, we show a comparison of positive rates across different thresholds for each sensitive group (fig. \ref{fig:PR}).  We further demonstrate the performance of the fairness metric (SPD) across the complete range of selected thresholds for each model (fig. \ref{fig:SPD}).

\begin{table}[]
\caption{Fairness metric (SPD) and model performance metric (AUC-ROC) for Adult data set. We report the average and the standard deviation across 10 runs and 101 thresholds.}
  \label{adult-table}
  \centering
  \small
  \resizebox{0.7\textwidth}{!}{%
\begin{tabular}{@{}llllll@{}}
\toprule
\multicolumn{1}{l}{\textbf{Metric}} & \textbf{Training data} & \multicolumn{1}{l}{\textbf{CatBoost}} & \multicolumn{1}{l}{\textbf{LightGBM}} & \multicolumn{1}{l}{\textbf{RandomForest}} & \multicolumn{1}{l}{\textbf{XGBoost}} \\ \midrule
\multicolumn{1}{c}{\multirow{6}{*}{\rotatebox[origin=c]{90}{SPD}}} & original& $0.194\pm0.117$  & $0.197\pm0.116$  & $0.197\pm0.096$& $0.198\pm0.117$ \\ \cmidrule(l){2-6}
  & synfair\_0.0  & $0.197\pm0.121$  & $0.198\pm0.122$  & $0.195\pm0.098$& $0.198\pm0.122$ \\ \cmidrule(l){2-6}
  & synfair\_0.25 & $0.144\pm0.082$  & $0.145\pm0.078$  & $0.141\pm0.053$& $0.144\pm0.078$ \\ \cmidrule(l){2-6}
  & synfair\_0.5  & $0.091\pm0.045$  & $0.092\pm0.043$  & $0.089\pm0.028$& $0.094\pm0.043$ \\ \cmidrule(l){2-6}
  & synfair\_0.75 & $0.045\pm0.026$&$0.051\pm0.03$&$0.045\pm0.016$&$0.049\pm0.025$ \\ \cmidrule(l){2-6}
  & synfair\_1.0  & $0.012\pm0.01$&$0.015\pm0.011$&$0.019\pm0.009$&$0.013\pm0.008$ \\\midrule
\multicolumn{1}{c}{\multirow{6}{*}{\rotatebox[origin=c]{90}{AUC}}} & original& $0.928$ & $0.927$ & $0.897$  & $0.929$\\ \cmidrule(l){2-6}
\multicolumn{1}{c}{}& synfair\_0.0  & $0.921\pm0.001$  & $0.920\pm0.002$& $0.889\pm0.002$& $0.920\pm0.002$  \\ \cmidrule(l){2-6}
\multicolumn{1}{c}{}& synfair\_0.25 & $0.918\pm0.002$  & $0.917\pm0.002$  & $0.881\pm0.003$& $0.917\pm0.002$ \\ \cmidrule(l){2-6}
\multicolumn{1}{c}{}& synfair\_0.5  & $0.912\pm0.002$  & $0.910\pm0.002$& $0.873\pm0.003$& $0.911\pm0.003$ \\ \cmidrule(l){2-6}
\multicolumn{1}{c}{}& synfair\_0.75 & $0.902\pm0.002$  & $0.901\pm0.002$  & $0.863\pm0.003$& $0.901\pm0.002$ \\ \cmidrule(l){2-6}
\multicolumn{1}{c}{}& synfair\_1.0  & $0.890\pm0.003$& $0.889\pm0.003$  & $0.853\pm0.003$& $0.889\pm0.003$ \\ \bottomrule
\end{tabular}%
}
\end{table}

\begin{table}[]
\caption{Fairness metric (SPD) and model performance metric (AUC-ROC) for Dutch census data set. We report the average and the standard deviation across 10 runs and 101 thresholds.}
  \label{dutch-table}
  \centering
  \small
  \resizebox{0.7\textwidth}{!}{%
\begin{tabular}{@{}llllll@{}}
\toprule
\multicolumn{1}{l}{\textbf{Metric}} & \textbf{Training data} & \multicolumn{1}{l}{\textbf{CatBoost}} & \multicolumn{1}{l}{\textbf{LightGBM}} & \multicolumn{1}{l}{\textbf{RandomForest}} & \multicolumn{1}{l}{\textbf{XGBoost}} \\ \midrule
\multicolumn{1}{c}{\multirow{6}{*}{\rotatebox[origin=c]{90}{SPD}}} & original& $0.292\pm0.133$  & $0.288\pm0.137$  & $0.293\pm0.092$& $0.289\pm0.140$  \\ \cmidrule(l){2-6}
& synfair\_0.0  & $0.293\pm0.138$  & $0.291\pm0.137$  & $0.292\pm0.095$& $0.291\pm0.139$ \\ \cmidrule(l){2-6}
& synfair\_0.25 & $0.217\pm0.112$  & $0.214\pm0.112$  & $0.213\pm0.073$& $0.215\pm0.115$ \\ \cmidrule(l){2-6}
& synfair\_0.5  & $0.143\pm0.078$  & $0.140\pm0.081$& $0.143\pm0.057$& $0.142\pm0.079$ \\ \cmidrule(l){2-6}
& synfair\_0.75 & $0.068\pm0.046$  & $0.067\pm0.047$  & $0.071\pm0.038$& $0.068\pm0.046$ \\ \cmidrule(l){2-6}
& synfair\_1.0 & $0.020\pm0.014$&$0.019\pm0.011$&$0.009\pm0.006$&$0.018\pm0.011$\\ \midrule
\multicolumn{1}{c}{\multirow{6}{*}{\rotatebox[origin=c]{90}{AUC}}} & original& $0.914$ & $0.914$ & $0.892$  & $0.913$\\ \cmidrule(l){2-6}
& synfair\_0.0  & $0.911\pm0.000$ & $0.911\pm0.001$  & $0.888\pm0.002$& $0.911\pm0.001$ \\ \cmidrule(l){2-6}
& synfair\_0.25 & $0.909\pm0.001$  & $0.909\pm0.001$  & $0.881\pm0.002$& $0.909\pm0.001$ \\ \cmidrule(l){2-6}
& synfair\_0.5  & $0.901\pm0.001$  & $0.901\pm0.001$  & $0.871\pm0.001$& $0.901\pm0.001$ \\ \cmidrule(l){2-6}
& synfair\_0.75 & $0.886\pm0.002$  & $0.885\pm0.002$  & $0.855\pm0.002$& $0.885\pm0.002$ \\ \cmidrule(l){2-6}
& synfair\_1.0  & $0.863\pm0.002$  & $0.864\pm0.002$  & $0.831\pm0.003$& $0.862\pm0.002$ \\ \bottomrule
\end{tabular}%
}
\end{table}

\section{Conclusion}
In this study, our primary goal is to investigate the potential of our approach in generating synthetic data that not only achieves Fairness by Design but also maintains a high degree of flexibility throughout the synthesis process. By Fairness by Design, we refer to the ability of a model trained on fair synthetic data, specifically with respect to statistical parity, to consistently produce fair predictions across various decision thresholds, even when inferring from real-world, biased data. Flexibility, in our context, signifies the absence of any preexisting assumptions or information requirements about the input data before training the synthetic data generator and the absence of the necessity for re-training the synthetic data generator while adjusting the fairness-utility tradeoff.\\
Our empirical results demonstrate that our approach effectively achieves these objectives. Across the two data sets subject to our analysis, we are able to reduce the statistical parity difference to levels well below 0.1 while preserving the desired flexibility. Moreover, even when employing the most stringent fairness corrections $\lambda=1$, the performance of downstream models, as measured by the AUC-ROC metric, exhibited a maximum decline of merely 5 percentage points.\\
Our approach involves the strategy of relocating the fairness correction to the post-processing stage of the synthetic data synthesis. We adapt and extend a method from the fairness literature to the generation of fair synthetic data. Specifically, our methodology ensures that conditional probabilities for the target column within unprivileged groups closely align with those of privileged groups within the synthetic population.\\
Looking ahead, this work sets the stage for the exploration of post-processing fairness strategies beyond statistical parity for providing Fairness by Design through fair synthetic data.

\newpage
\bibliographystyle{unsrt}
\bibliography{references}

\begin{thebibliography}{10}

\bibitem{fairsummary}
Suvodeep Majumder, Joymallya Chakraborty, Gina~R. Bai, Kathryn~T. Stolee, and Tim Menzies.
\newblock Fair enough: Searching for sufficient measures of fairness, 2022.

\bibitem{fairgen}
Bhushan Chaudhari, Himanshu Chaudhary, Aakash Agarwal, Kamna Meena, and Tanmoy Bhowmik.
\newblock Fairgen: Fair synthetic data generation, 2022.

\bibitem{fairgan}
Depeng Xu, Shuhan Yuan, Lu~Zhang, and Xintao Wu.
\newblock Fairgan: Fairness-aware generative adversarial networks, 2018.

\bibitem{tabfairgan}
Amirarsalan Rajabi and Ozlem~Ozmen Garibay.
\newblock Tabfairgan: Fair tabular data generation with generative adversarial networks, 2021.

\bibitem{ourfairness}
Paul Tiwald, Alexandra Ebert, and Daniel~T. Soukup.
\newblock Representative \& fair synthetic data, 2021.

\bibitem{notnecessarily}
Yam Eitan, Nathan Cavaglione, Michael Arbel, and Samuel Cohen.
\newblock Fair synthetic data does not necessarily lead to fair models.
\newblock In {\em NeurIPS 2022 Workshop on Synthetic Data for Empowering ML Research}, 2022.

\bibitem{decaf}
Boris van Breugel, Trent Kyono, Jeroen Berrevoets, and Mihaela van~der Schaar.
\newblock Decaf: Generating fair synthetic data using causally-aware generative networks, 2021.

\bibitem{prefair}
David Pujol, Amir Gilad, and Ashwin Machanavajjhala.
\newblock Prefair: Privately generating justifiably fair synthetic data, 2023.

\bibitem{wasserstein}
Ray Jiang, Aldo Pacchiano, Tom Stepleton, Heinrich Jiang, and Silvia Chiappa.
\newblock Wasserstein fair classification, 2019.

\bibitem{georepair}
Kweku Kwegyir-Aggrey, Jessica Dai, A.~Feder Cooper, John Dickerson, and Keegan Hines.
\newblock Geometric repair for fair classification at any decision threshold, 2023.

\bibitem{quantileregression}
Meichen Liu, Lei Ding, Dengdeng Yu, Wulong Liu, Linglong Kong, and Bei Jiang.
\newblock Conformalized fairness via quantile regression, 2022.

\bibitem{fairregression}
Evgenii Chzhen, Christophe Denis, Mohamed Hebiri, Luca Oneto, and Massimiliano Pontil.
\newblock Fair regression with wasserstein barycenters, 2020.

\bibitem{wassersteindistance}
L.~V. Kantorovich.
\newblock Mathematical methods of organizing and planning production.
\newblock {\em Management Science}, 6(4):366--422, 1960.

\bibitem{projectionfairness}
Thibaut~Le Gouic, Jean-Michel Loubes, and Philippe Rigollet.
\newblock Projection to fairness in statistical learning, 2020.

\bibitem{mostly}
Mostly-ai synthetic data platform.
\newblock \url{https://mostly.ai/synthetic-data-platform/generate-synthetic-data/}, 2022.

\bibitem{sampling}
Ari Holtzman, Jan Buys, Li~Du, Maxwell Forbes, and Yejin Choi.
\newblock The curious case of neural text degeneration, 2020.

\bibitem{ds_survey}
Tai~Le Quy, Arjun Roy, Vasileios Iosifidis, Wenbin Zhang, and Eirini Ntoutsi.
\newblock A survey on datasets for fairness-aware machine learning.
\newblock {\em {WIREs} Data Mining and Knowledge Discovery}, 12(3), mar 2022.

\bibitem{adult_ds}
Barry Becker and Ronny Kohavi.
\newblock {Adult}.
\newblock UCI Machine Learning Repository, 1996.
\newblock {DOI}: https://doi.org/10.24432/C5XW20.

\bibitem{dutch_ds}
Paul Van~der Laan.
\newblock {\em The 2001 Census in the Netherlands: Integration of Registers and Surveys}, pages 39--52.
\newblock 12 2001.

\bibitem{autogluon}
Rasool Fakoor, Jonas~W Mueller, Nick Erickson, Pratik Chaudhari, and Alexander~J Smola.
\newblock Fast, accurate, and simple models for tabular data via augmented distillation.
\newblock {\em Advances in Neural Information Processing Systems}, 33, 2020.

\end{thebibliography}








\end{document}